\documentclass[lettersize,journal]{IEEEtran}
\usepackage[T1]{fontenc}
\usepackage{amsmath,amsfonts}
\usepackage{algorithmic}
\usepackage{algorithm}
\usepackage{array}
\usepackage[caption=false,font=normalsize,labelfont=sf,textfont=sf]{subfig}
\usepackage{textcomp}
\usepackage{stfloats}
\usepackage{url}
\usepackage{verbatim}
\usepackage{graphicx}
\usepackage{cite}
\usepackage{booktabs}
\usepackage{multirow}

\begin{document}

\newcommand{\method}{FMA-Net}

\title{FMA-Net: Flexible Modality Adaptation for Robust Multimodal \\ Face Presentation Attack Detection}

\author{Yuhao Gao, Yichen Shi, ~\IEEEmembership{Member,~IEEE},
    Zitong Yu,~\IEEEmembership{Senior~Member,~IEEE},
    Hongyang Wang,~\IEEEmembership{Member,~IEEE},
    Hongrui Li,~\IEEEmembership{Member,~IEEE}, Ruixin Zhang, Shouhong Ding,
    Jun Feng

    \thanks{This work was sponsored by CCF-Tencent Open Fund, Natural Science Foundation of  Hebei Province (Grant No. F2024210005) and National Natural Science Foundation of China (Grant No. 62576076). Yuhao Gao, Yichen Shi, and Zitong Yu are co-first authors and contribute equally. (Corresponding author: Jun Feng)}

    \thanks{Y. Gao, H. Li and J. Feng are with School of Information Science and Technology, Shijiazhuang Tiedao University, Shijiazhuang, China. Email: \{1202210044, 1202410034\}@student.stdu.edu.cn, fengjun@stdu.edu.cn}

    \thanks{H. Wang is with the School of Integrated Circuits and Micro-Nano Electronics, Fudan University, Shanghai, China. Email: 26113040025@m.fudan.edu.cn}

    \thanks{Y. Shi is with School of Integrated Circuit, Shanghai Jiao Tong University, Shanghai, China, and also with Eastern Institute of Technology, Ningbo, Zhejiang, China. Email: shiyichen@sjtu.edu.cn}

    \thanks{Z. Yu is with School of Computing and Information Technology, Great Bay University, Dongguan, China. E-mail: yuzitong@gbu.edu.cn}

    \thanks{R. Zhang and S. Ding are with Tencent Youtu Lab, Shanghai, China. E-mail: \{ruixinzhang, ericshding\}@tencent.com}
}

\markboth{FMA-Net: arXiv Preprint}
{Gao \MakeLowercase{\textit{et al.}}: FMA-Net: Flexible Modality Adaptation for Robust Multimodal Face Presentation Attack Detection}

\maketitle

\begin{abstract}
Face Presentation Attack Detection (fPAD) is a fundamental component of secure face recognition systems. While multimodal fPAD improves robustness by exploiting complementary cues from RGB, depth, and infrared (IR) modalities, its practical deployment remains challenging when one or more modalities are unavailable or incomplete. Existing flexible-modality fPAD methods address this issue mainly through zero-filling or modality omission, which may discard useful cross-modal information and introduce modality-dependent bias. To overcome these limitations, we propose the \textbf{Flexible Modality Adaptation Network (FMA-Net)}, a unified framework for flexible-modality fPAD that compensates for missing modalities through instruction-guided latent-space translation. Specifically, the Face Modality Instruct Translator (FMIT) estimates target modality observations from available RGB inputs, while a context-aware category enhancement mechanism embeds learnable text prompts as modality-invariant semantic anchors to alleviate modality bias. The Facial Feature Adapter (\textbf{F\textsuperscript{2}A}) module further enhances discriminative facial representations by integrating global facial structure and local spoof-relevant cues. In addition, a Synthetic-to-Real Face Domain Alignment (SR-FDA) module reduces the feature discrepancy between translated and real modalities, with benefits that depend on the missing-modality pattern and data domain. Extensive experiments on the WMCA and PADISI-Face datasets validate the effectiveness of the proposed framework. On WMCA under intra-domain evaluation with a 70\% modality missing ratio, FMA-Net achieves a mean ACER of 3.22\% across four missing-modality protocols, compared with 4.24\% for the strongest baseline by the same mean criterion. When both depth and IR are entirely unavailable, it achieves an ACER of 5.23\%, compared with 5.99\% for the best competing result in that setting. Cross-domain experiments further reveal the limitations of synthetic-to-real alignment under dataset shifts. These results demonstrate that FMA-Net provides a practical solution for fPAD under flexible and incomplete modality scenarios.

\end{abstract}


\begin{figure}[!t]
\centering
\includegraphics[width=\linewidth]{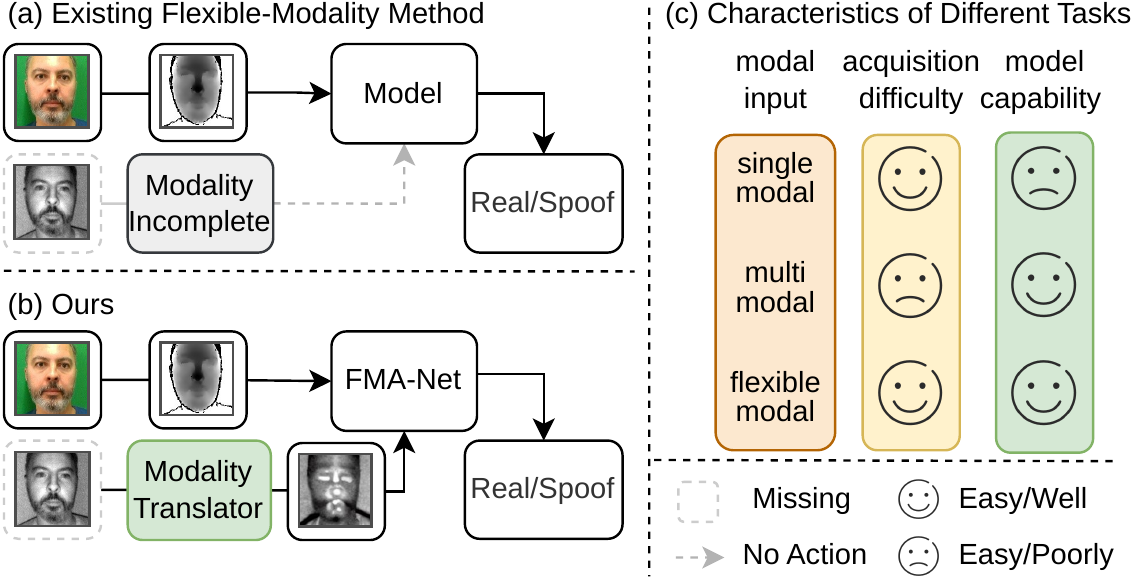}
\caption{Comparison between our method and existing flexible-modality face Presentation Attack Detection (fPAD) approaches under a missing infrared (IR) modality.
(a) Existing methods address missing modalities via skipping or zero-padding.
(b) FMA-Net employs a modality-translation module to reconstruct the missing modality, enabling all modalities to be used by the fPAD model for robust inference.
(c) Comparative analysis of task settings: unimodal, multimodal, and flexible-modality, in terms of input-modality structure, data-acquisition complexity, and model capabilities. The source face images are from the WMCA dataset~\cite{george2019biometric}.}
\label{motivation}
\end{figure}

\section{Introduction}
\label{sec:Introduction}
With the rapid advancement of face recognition technologies, face presentation attack detection (fPAD) has become a crucial component for secure identity verification. From smartphone unlocking to facial payment systems, face recognition is increasingly integrated into daily life. However, as a sensitive biometric trait, facial data is vulnerable to theft and misuse. Attackers can readily exploit publicly available facial images to launch presentation attacks by displaying photographs, replaying videos, or fabricating 3D masks. Consequently, fPAD aims to determine whether the presented face originates from a live subject rather than a spoofed medium, and has attracted extensive research attention, particularly in high-security applications.

Recent advances in fPAD highlight the effectiveness of multimodal approaches, which leverage complementary information from heterogeneous modalities such as RGB, depth, IR, and thermal. By combining diverse cues, these methods significantly enhance robustness against sophisticated attacks including high-quality replays and 3D masks. However, their reliance on specialized hardware and fixed modality configurations limits scalability and practical adoption. Prior work on flexible-modality fPAD~\cite{yu2023flexible} seeks unified models that handle arbitrary modality combinations. Yet, existing strategies for coping with missing modalities---such as zero-filling or direct omission~\cite{xiao2024}---inevitably discard valuable information and lead to suboptimal performance. This underscores a critical gap: how to effectively compensate for unavailable modalities while preserving the benefits of multimodal fusion. Motivated by this challenge, this work investigates modality translation as a flexible compensation strategy for incomplete-modality fPAD, aiming to exploit complementary modality cues without requiring all modalities to be physically available at inference.

To address this problem, we propose the \textbf{Flexible Modality Adaptation Network (FMA-Net)}, a unified framework for robust fPAD under incomplete modality availability. It first employs the \textbf{Face Modality Instruct Translator (FMIT)} to estimate target modality observations from RGB inputs through instruction-guided latent-space translation. Then, a lightweight \textbf{Facial Feature Adapter (F\textsuperscript{2}A)} enhances face-specific representation learning by integrating global facial structure and fine-grained local spoof-relevant cues. Finally, to mitigate the discrepancy between translated and real modalities, \textbf{Synthetic-to-Real Face Domain Alignment (SR-FDA)} performs adversarial feature alignment to address modality-origin differences during detector training; its benefit under cross-domain evaluation is assessed separately. To comprehensively evaluate flexible-modality fPAD under practical deployment conditions, we further establish a \textbf{flexible-modality evaluation protocol} that covers both intra-domain and cross-domain settings and systematically simulates partial- and full-modality missing scenarios. The main contributions of this work are summarized as follows:

\begin{itemize}
    \item \textbf{Unified fPAD under Incomplete Modalities.}
    We develop a single training--inference paradigm that directly uses available modalities and compensates for unavailable ones within the same pipeline, thereby reducing dependence on fixed modality configurations and improving robustness across varying missing ratios and modality combinations.

    \item \textbf{Face Modality Instruct Translator (FMIT).}
    We design an instruction-guided latent diffusion engine that translates RGB inputs into target modality observations, such as depth and IR, to support both training-time modality augmentation and inference-time missing-modality compensation.

    \item \textbf{Facial Feature Adapter (F\textsuperscript{2}A).}
    We introduce a lightweight face-focused adapter that enhances facial representation learning through global-structure modeling and local feature aggregation, improving sensitivity to spoof-relevant cues under both real and translated modality inputs.

    \item \textbf{Synthetic-to-Real Face Domain Alignment (SR-FDA).}
    We perform adversarial feature alignment between translated and real modalities to reduce translation-induced discrepancies during detector training, and evaluate both its benefits and its limitations across intra-domain and cross-domain settings.

\end{itemize}

\section{Related Work}
\subsection{Face Presentation Attack Detection}

Depending on modality availability and modeling strategy, existing multimodal fPAD methods can be broadly categorized into two paradigms: fixed multimodal fusion and flexible-modality learning.

With the rapid advancement of sensing technologies, multimodal methods have become increasingly prominent in face presentation attack detection (fPAD). These methods exploit complementary appearance, geometric, and spectral cues from RGB, depth, IR, and other modalities. Representative works include the multi-channel CNN with the WMCA dataset~\cite{george2019biometric} and the PADISI-Face dataset with continual learning~\cite{rostami2021detection}. Advanced designs integrate modality-parallel branches~\cite{yu2020multi}, cross-modality losses~\cite{george2021cross}, dual-stream attention fusion~\cite{deng2023attention}, hybrid structures~\cite{wang2022conv}, and hyperbolic embeddings~\cite{han2023hyperbolic}. GAN-based cross-modality translation has also been explored to synthesize missing modality observations~\cite{liu2021face}, but such methods usually rely on task-specific translation mappings and do not explicitly address the translated-real modality discrepancy during flexible-modality inference.

However, despite their strong performance, multimodal methods typically assume full-modality availability during both training and deployment, which is often unrealistic in practice. This limitation has motivated research into flexible-modality frameworks that can adapt to varying input availability and device heterogeneity. For example, Liu et al.~\cite{liu2021face} employed GAN-based RGB-to-depth translation to improve cross-modality generalization, while FM-ViT and FM-CLIP~\cite{liu2023fm,liu2024fm} introduced flexible-modality representations based on Transformer and vision-language modeling. Other studies introduced input-aware routing mechanisms to dynamically adjust architectures based on modality presence. These methods improve fPAD applicability in edge computing and resource-constrained environments. Different from these discriminative flexible-modality frameworks, our work uses instruction-guided modality translation to provide auxiliary modality observations and further aligns translated and real modality features for fPAD.

Early fPAD systems primarily relied on convolutional neural networks (CNNs) with multi-branch or attention-based fusion and cross-modal supervision~\cite{yu2020multi,george2021cross,deng2023attention}. Vision Transformers (ViTs), often instantiated as Conv--Transformer hybrids, were subsequently introduced to capture long-range dependencies and token-level relationships~\cite{wang2022conv}. Building upon these discriminative architectures, large model-driven approaches have emerged, particularly those employing vision-language models (VLMs) such as CLIP~\cite{radford2021learning}. By aligning visual and textual representations via contrastive learning, these models support natural language-guided detection. Recent works demonstrated the effectiveness of linguistic prompts for unsupervised and open-set fPAD~\cite{fang2024vl},~\cite{liu2024cfpl}. With the rise of multimodal large language models (MLLMs), efforts have been made to establish benchmarks~\cite{shi2025shield} and propose interpretable frameworks integrating language reasoning with multimodal representation learning~\cite{zhang2025interpretable}. These advancements enhance semantic understanding, generalization, and interpretability, paving the way for language-aware security systems. However, most language-aware fPAD methods focus on semantic supervision or open-set recognition, while the problem of compensating unavailable physical modalities such as depth and IR remains underexplored.

Despite these advances, most existing approaches still suffer from limitations when facing missing modalities. Multimodal systems often assume full-modality availability during both training and inference, which rarely holds in real-world scenarios. Flexible-modality strategies alleviate this issue to some extent, but many of them rely on omission, zero-filling, or discriminative adaptation, leaving the complementary cues of unavailable modalities insufficiently exploited. This results in underutilization of modality-specific knowledge and may hamper system robustness. To address this gap, our work introduces an instruction-guided modality compensation framework that estimates auxiliary modality observations and incorporates synthetic-to-real feature alignment, aiming to improve anti-spoofing robustness under incomplete inputs.

\subsection{Generation Model For fPAD}
Early deep generative models---including GANs~\cite{goodfellow2020generative}, VAEs~\cite{kingma2013auto}, and autoregressive models~\cite{van2016conditional}---substantially improved image synthesis quality and enabled image-to-image translation~\cite{zhu2017unpaired}, but typically faced trade-offs among fidelity, stability, and efficiency. Recent diffusion models establish a more stable training paradigm via a forward noising and reverse denoising process, from DDPM~\cite{ho2020denoising} to accelerated sampling in DDIM~\cite{song2020denoising}. Latent diffusion further compresses the process into a learned latent space~\cite{rombach2022high}, making high-quality conditional generation computationally feasible. Building on this, controllable diffusion (e.g., ControlNet~\cite{zhang2023adding}) injects structure priors to enhance conditional precision. These properties make latent diffusion suitable for modality translation, where the target modality may contain ambiguous or one-to-many mappings conditioned on RGB observations.

In face restoration, RestoreFormer by Wang et al.~\cite{wang2022restoreformer} uses cross-attention and a high-quality feature dictionary to recover facial details. Diffusion-based approaches have also shown promise. For example, Qiu et al.~\cite{qiu2023diffbfr} introduced DiffBFR, which leverages diffusion-based models to recover high-quality facial images while avoiding the training instability commonly associated with GANs. In the domain of facial security, Liu et al.~\cite{liu2024adv} proposed Adv-Diffusion, which utilizes latent diffusion models to generate high-fidelity adversarial examples capable of deceiving facial recognition systems, highlighting the utility of diffusion models in adversarial attack tasks. For style transfer, Brooks et al.~\cite{brooks2023instructpix2pix} developed InstructPix2Pix, enabling expression and style editing through text prompts, while Qi et al.~\cite{qi2024deadiff} proposed DEADiff, which enhances style controllability in text-to-image generation via disentangled representations. Taken together, these studies underscore the strong expressive and controllable capabilities of diffusion models in facial image restoration, security, and style manipulation, providing a technical foundation for instruction-guided face modality translation. Nevertheless, modality translation for fPAD differs from general facial image restoration or editing because the translated observations should preserve spoof-relevant cues rather than merely improve perceptual quality.

\begin{figure*}[htbp]
    \centering
    \includegraphics[width=1\linewidth]{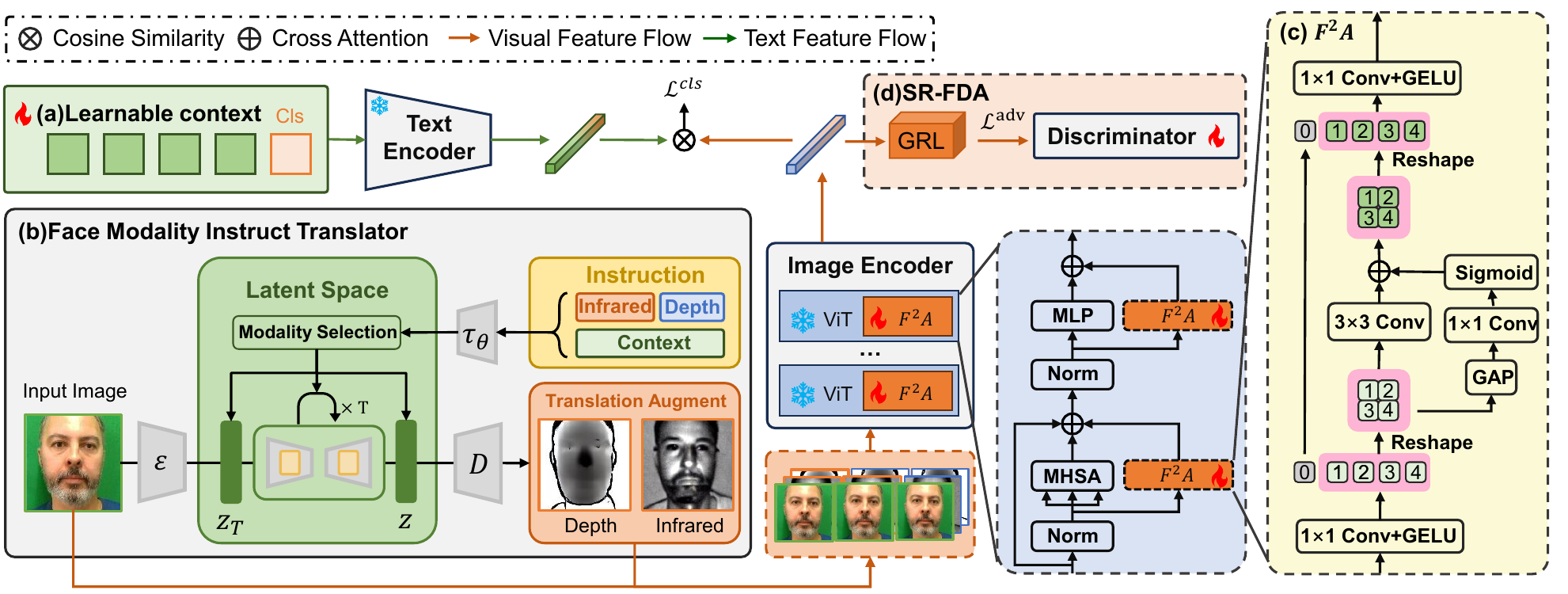}
    \caption{Overview of the proposed FMA-Net. (a) Learnable context and the text encoder capture face-related semantic features. (b) FMIT estimates missing modality observations in the latent space. (c) The F\textsuperscript{2}A module extracts visual features from both real and translated multimodal inputs. (d) SR-FDA mitigates domain shift by encouraging domain-invariant feature learning. The source face images are from the WMCA dataset~\cite{george2019biometric}.}
    \label{fig:pipeline}
\end{figure*}

\label{sec:Literature}

\section{The Proposed Method}
\label{sec:Proposed Method}
This section presents the \textbf{Flexible Modality Adaptation Network (FMA-Net)}, a unified framework for flexible-modality fPAD that directly uses available modalities and compensates for unavailable ones through translated auxiliary observations. We first outline the overall pipeline in Sec.~\ref{sec:pipeline}. We then introduce the \textbf{Face Modality Instruct Translator (FMIT)}, an instruction-guided latent-diffusion translator that estimates target modality observations, such as depth and IR, from RGB inputs, as described in Sec.~\ref{sec:Diffusion_modality_translation}. Next, we describe learnable prompts together with the \textbf{Facial Feature Adapter (F\textsuperscript{2}A)}, which aggregates multi-scale and region-aware facial cues to strengthen face-specific representation learning, as described in Sec.~\ref{sec:FAS_module}. Finally, we detail \textbf{Synthetic-to-Real Face Domain Alignment (SR-FDA)}, which reduces the distribution gap between translated and real modalities during detector training, with protocol-dependent effects evaluated in the experiments, as described in Sec.~\ref{sec:DG_module}.

\subsection{Overview of FMA-Net}
\label{sec:pipeline}

To handle incomplete modality acquisition in practical deployment, we develop a unified flexible-modality fPAD framework that integrates modality translation, discriminative feature adaptation, and synthetic-to-real domain alignment into a cohesive pipeline, as illustrated in Fig.~\ref{fig:pipeline}. The framework first employs the \textbf{Face Modality Instruct Translator (FMIT)} to estimate auxiliary depth or IR observations from accessible RGB inputs through instruction-guided latent diffusion. The translated and real modalities are then jointly processed by an image encoder equipped with the \textbf{Facial Feature Adapter (F\textsuperscript{2}A)}, which enhances spoof-discriminative representation via adaptive local--global feature aggregation. To mitigate the distribution gap between translated and real modalities, the \textbf{Synthetic-to-Real Face Domain Alignment (SR-FDA)} module is incorporated based on adversarial learning. Through this joint design, the proposed system improves robustness and generalization under varying modality availability. FMIT is trained separately and kept frozen during detector training. The learnable detector prompts, facial feature adapter, and SR-FDA domain discriminator are optimized jointly during detector training; the detector objectives do not update FMIT. Each component is detailed in the following subsections.

\subsection{Face Modality Instruct Translator}
\label{sec:Diffusion_modality_translation}

\begin{figure}[htbp]
    \centering
    \includegraphics[width=\linewidth]{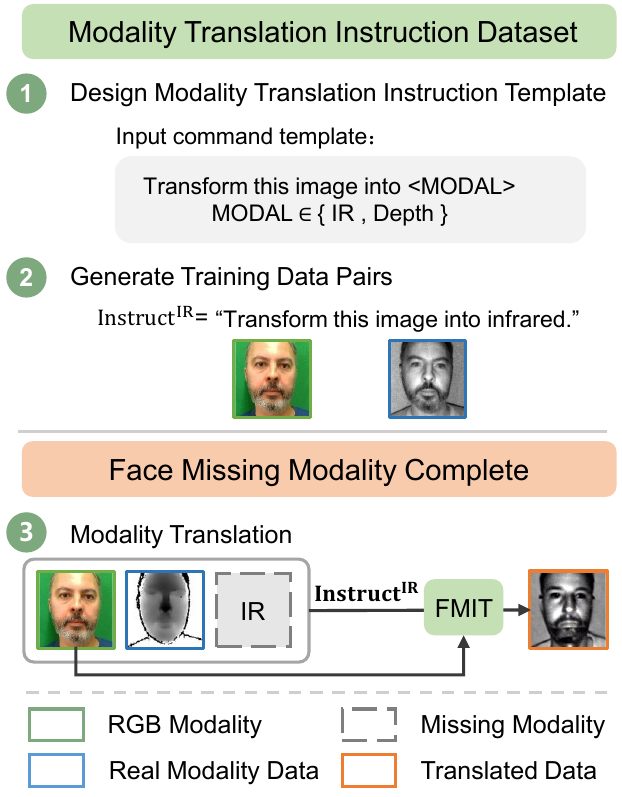 }
    \caption{Overview of the modality translation instruction dataset construction and the face missing-modality compensation process. Steps 1--2 illustrate dataset creation through instruction design and paired data generation, while step 3 shows modality translation by FMIT for estimating missing modality observations. The source face images are from the WMCA dataset~\cite{george2019biometric}.}
    \label{fig:generate_dataset_building}
\end{figure}

In flexible-modality fPAD, different modalities exhibit complementary appearance, geometric, and spectral cues for distinguishing bona fide and attack presentations. However, in practical applications, modality information is frequently incomplete, especially under flexible-modality protocols where not all samples contain full multimodal data. Existing modality-robust learning or prompt-learning methods alleviate this issue to some extent, but they do not explicitly estimate the unavailable modality observations that may provide complementary spoof-relevant cues.

We propose the \textbf{Face Modality Instruct Translator (FMIT)}, which takes a visible-light image and a modality translation instruction as inputs and estimates target modality observations such as IR and depth. The task is formulated as supervised instruction-guided translation: the instruction generator produces a concise prompt, such as ``Transform this image into the X modality'', according to the target modality, and the translator learns to produce the corresponding target-modality observation. To train FMIT, a paired instruction dataset is constructed by extracting RGB--IR and RGB--depth samples from a multimodal face database, as illustrated in Fig.~\ref{fig:generate_dataset_building}. Each training triplet consists of a visible-light image, a modality instruction, and the paired target-modality image. The instructions specify only the target modality and do not require the bona fide/spoof label or attack category, which are unavailable at inference time. This permits the same instruction template to be used for both seen and unseen attack categories.

The input data include a visible-light image and any available auxiliary modalities, such as IR or depth images. Based on the modality availability of the input sample, the instruction generator produces a corresponding set of modality translation instructions, denoted as $P$. For instance, if the IR modality is unavailable, the generated instruction $P_{\mathrm{IR}}$ guides FMIT to estimate an IR observation from the RGB input.

FMIT performs modality compensation based on a stable diffusion framework, with the core translation function denoted as $f(\cdot)$. When the target modality is IR, the input consists of the RGB image $X_{\mathrm{RGB}}$ and the translation instruction $P_{\mathrm{IR}}$. The translated observation $\hat{X}_{\mathrm{IR}}$ is computed as:

    \begin{equation}
    \label{deqn_ex9}
    \hat{X}_{\text{IR}} = f \left( X_{\text{RGB}}, P_{\text{IR}} \right)
    \end{equation}

FMIT performs conditional modality translation in the latent space of a pretrained VAE. For a paired RGB observation $X_{\mathrm{RGB}}$ and target-modality observation $X_m$, where $m\in\{\mathrm{IR},\mathrm{Depth}\}$, let $z_0=\mathcal{E}(X_m)$ denote the clean target latent and $c_{\mathrm{RGB}}=\mathcal{E}(X_{\mathrm{RGB}})$ the image condition. The VAE decoder $\mathcal{D}$ maps the predicted target latent back to the target-modality image.

During training, Gaussian noise is added to the target latent according to the diffusion noise schedule:
\begin{equation}
 z_t=a_t z_0+\sigma_t\epsilon,\qquad \epsilon\sim\mathcal{N}(0,I),\quad t\in\{1,\ldots,T\}.
\end{equation}
Here, $a_t$ and $\sigma_t$ are the time-dependent signal and noise coefficients, with $a_t^2+\sigma_t^2=1$. These coefficients are distinct from the modality-missing probability $\alpha$ used in the experiments. The clean latent $z_0$ is obtained from the target observation; only the added noise $\epsilon$ is sampled from the standard normal distribution.

The instruction encoder $\tau_\theta$ maps the target-modality instruction $p$ to a text condition. The denoising UNet $\epsilon_\theta$ receives the noisy target latent, timestep, RGB condition, and text condition. Its conditional noise-prediction objective is
\begin{equation}
\begin{aligned}
L_{\mathrm{cond}}={}&\mathbb{E}_{(X_{\mathrm{RGB}},X_m,p),\,t,\,\epsilon}
\big[\|\epsilon-\epsilon_\theta(z_t,t,c_{\mathrm{RGB}},\tau_\theta(p))\|_2^2\big].
\end{aligned}
\end{equation}
The paired observations and instructions are sampled from the modality-translation training data. The RGB condition supplies image content, while the instruction identifies the target modality. The pretrained VAE and instruction encoder are frozen, and only the denoising UNet is fine-tuned in this stage; the retained encoder notation $\tau_\theta$ does not imply parameter sharing with the UNet.

By introducing conditional control, the model can perform instruction-guided translation across different facial modalities and estimate missing modality observations, thereby providing auxiliary inputs for downstream flexible-modality fPAD.

\subsection{Learnable Prompt Learning For fPAD}
\label{sec:FAS_module}

Feature extraction is conducted through an image encoder $E_{\mathrm{I}}$ and a text encoder $E_{\mathrm{T}}$. In the textual stream, a context-aware class enhancement strategy is employed to improve cross-modality alignment of textual features and alleviate modality bias in flexible-modality scenarios. In the visual stream, we introduce the \textbf{Facial Feature Adapter (F\textsuperscript{2}A)} into $E_{\mathrm{I}}$ to enhance face-specific representation learning. Unlike a conventional adapter that mainly performs generic feature transformation or parameter-efficient tuning, F\textsuperscript{2}A is designed for fPAD by jointly modeling global facial structure and local spoof-relevant cues from both real and translated modality inputs.

As illustrated in Fig.~\ref{fig:pipeline}, the input facial images may include real RGB observations together with real or translated depth and IR observations, depending on the modality availability of each sample. During each fPAD process, the model processes textual and visual feature streams in parallel.

\subsubsection{Text Feature Flow}

For the textual stream, the model uses the class label set $Labels=\{\mathrm{Bonafide},\mathrm{Spoof}\}$ together with contextual prompts. Let $v_i$ denote the context prompt containing the name of class $i$. These prompts are processed by the text encoder $E_{\mathrm{T}}$ to generate textual features, which are represented as:

    \begin{equation}
    \label{deqn_ex1}
    \hat{w}_i = \frac{E_{\mathrm{T}}(v_i)}{\|E_{\mathrm{T}}(v_i)\|_2}, \quad i\in\{1,\ldots,K\}
    \end{equation}

To further enhance cross-modality alignment, Context Optimization (CoOp) is employed to learn modality-independent textual prompts. By replacing manually designed fixed templates with learnable contextual embeddings, CoOp improves the representation of class semantics and provides modality-invariant semantic anchors for flexible-modality fPAD.

For reference, the generic shared-context form concatenates context vectors with a class name as follows. Our experiments use class-specific context with the class token in the middle, as specified below.

    \begin{equation}
    \label{deqn_ex3}
    v = [V]_1 [V]_2 \cdots [V]_M [\textit{CLASS}]
    \end{equation}

Here, $[V]_m\ (m \in \{1,\ldots,M\})$ denotes the learnable context vectors, where $M$ represents the number of such vectors. The token ``[CLASS]'' serves as a class placeholder, indicating the textual description of the class label. $K$ is the number of categories. The normalized text feature $\hat{w}_i$ is defined in Eq.~\ref{deqn_ex1}. Let $F$ denote the sum of the normalized modality features, defined below. The score $\hat{w}_i^\top F$ is the sum of the per-modality cosine similarities; the summed feature is not normalized again. The probability that the input image belongs to class $i$ is computed using a positive classification temperature $T_{\mathrm{cls}}$. This scalar controls the sharpness of the class probabilities and is distinct from the FMIT instruction encoder $\tau_\theta$. The probability is given by:

    \begin{equation}
    \label{deqn_ex4}
    p(y = i \mid x) = \frac{\exp \left( \hat{w}_i^\top F / T_{\mathrm{cls}} \right)}{ \sum\limits_{j = 1}^{K} \exp \left( \hat{w}_j^\top F / T_{\mathrm{cls}} \right) }
    \end{equation}

To further enhance the flexibility of prompt representation, the class label can be inserted at different positions within the prompt. In addition to the aforementioned postfix structure, the class label can also be embedded in the middle of the context vectors. An alternative structure of the prompt vector is as follows:

    \begin{equation}
    \label{deqn_ex5}
    v = [V]_1 \cdots [V]_{\frac{M}{2}} [\textit{CLASS}] [V]_{\frac{M}{2} + 1} \cdots [V]_M
    \end{equation}

To enhance class-specific representations, a Class-Specific Context (CSC) strategy is introduced, where a unique set of context vectors is independently learned for each class. These vectors are optimized per class, instead of being shared across classes.

    \begin{equation}
    \begin{aligned}
    v_i &= [V]_1^i\cdots[V]_{M/2}^i[\mathrm{CLASS}_i][V]_{M/2+1}^i\cdots[V]_M^i
    \end{aligned}
    \end{equation}

The class-specific context vectors are independently parameterized and optimized by the classification objective. We use this CSC formulation with the middle class-token position in all reported detector experiments.

\subsubsection{Visual Feature Flow}

For the visual stream, the modality set $\{X_{\mathrm{RGB}}, X_{\mathrm{IR}}, X_{\mathrm{Depth}}\}$ may contain both real observations and translated auxiliary observations, depending on the missing-modality condition. Each modality input is processed by the image encoder equipped with F\textsuperscript{2}A, denoted $E_{\mathrm{I}}^{\mathrm{F^2A}}$, where the adapters operate inside the visual Transformer blocks rather than on raw input images. The extracted modality features are individually normalized and summed in the shared embedding space. Let $\mathcal{M}(x)$ contain the observed or completed modalities used for sample $x$; an uncompleted missing modality makes no contribution. Feature extraction and aggregation are given by:

    \begin{equation}
    \begin{aligned}
    F_j &= E_{\mathrm{I}}^{\mathrm{F^2A}}(X_j),
    j\in \{ \text{RGB}, \text{IR}, \text{Depth} \}
    \end{aligned}
    \end{equation}

    \begin{equation}
    \label{deqn_ex7}
    F = \sum_{j\in\mathcal{M}(x)} \frac{F_j}{\|F_j\|_2}
    \end{equation}

Since CLIP's image encoder $E_{\mathrm{I}}$ is pretrained on general-purpose image-text data, it may be less sensitive to fine-grained facial textures and local spoof artifacts that are critical for fPAD. To address this limitation, we propose the \textbf{Facial Feature Adapter (F\textsuperscript{2}A)}, a lightweight fPAD-oriented adaptation module inserted into the visual stream. Unlike conventional adapters that mainly learn generic residual transformations, F\textsuperscript{2}A explicitly combines global facial structure modeling with local spoof-cue aggregation, with the aim of strengthening facial feature extraction for fPAD. The module consists of four components:

\begin{itemize}
    \item \textbf{Feature Reduction (FR)}: A $1 \times 1$ convolution reduces the channel dimension of the input feature map while preserving spatial locations, producing a compact representation for subsequent global and local modeling.
    \item \textbf{Global Feature Extractor (GFE)}: Global average pooling (GAP), followed by a $1 \times 1$ convolution and a Sigmoid activation, generates recalibration weights that emphasize global facial structure and contour-level cues.
    \item \textbf{Local Feature Extractor (LFE)}: A $3 \times 3$ convolution captures local spatial patterns from facial regions such as the eyes, nose, and mask boundaries, improving sensitivity to fine-grained spoof-relevant cues.
    \item \textbf{Fusion Alignment (FA)}: The global and local branches interact through cross-attention. The fused representation is then projected back to the original dimension $D$ for downstream fPAD classification.
\end{itemize}

    \begin{equation}
    \begin{cases}
    F_G = GFE \left( FR(F) \right) \\
    F_L = LFE \left( FR(F) \right) \\
    F_a = FA(F_\mathrm{G}, F_\mathrm{L})
    \end{cases}
    \end{equation}

Here, $FR(\cdot)$ reduces channel dimensionality, $GFE(\cdot)$ captures global structural responses, $LFE(\cdot)$ extracts local spatial cues with a $3\times3$ convolution, and $FA(\cdot,\cdot)$ denotes the fusion-alignment module, which integrates the global and local features through cross-attention and projects the fused representation back to dimension $D$. This design differs from a conventional adapter by explicitly decomposing facial representation adaptation into global-structure and local-spoof-cue branches.

The class score is the inner product of the normalized text feature and the summed visual feature, equivalently the sum of the per-modality cosine similarities. The highest score determines the final detection result. The final predicted outcome $\hat{y}$ is given by:

    \begin{equation}
    \label{eq_9}
    \hat{y} = \arg\max_{i \in Labels} \, \hat{w}_i^\top F
    \end{equation}

\subsection{Synthetic-to-Real Face Domain Alignment}
\label{sec:DG_module}

Since translated auxiliary observations are produced by the face modality translator, they may introduce domain shift even when their visual appearance is close to that of real observations. This shift primarily appears in distributional characteristics such as style, noise, and texture statistics that are not directly related to fPAD decision making.

To mitigate the aforementioned issue, the \textbf{Synthetic-to-Real Face Domain Alignment (SR-FDA)} module is introduced on top of the flexible-modality translation-completion strategy, aiming to alleviate the adverse impact of domain shift caused by modality synthesis on detection performance. This module adopts adversarial learning to optimize the image feature extractor, encouraging it to preserve spoof-discriminative cues while suppressing modality-source-specific deviations.

Inspired by Domain-Adversarial Training of Neural Networks (DANN) \cite{ganin2016domain}, the input images are divided into two domains according to modality origin: the real-observation domain and the translated-observation domain. Each domain contains both bona fide and presentation attack samples. During training, SR-FDA jointly pursues two objectives: 1) optimizing the image feature extractor $E_{\mathrm{I}}$ and the classifier $\omega$ for accurate bona fide/spoof classification across both domains; and 2) introducing a domain discriminator $d$ to predict whether a visual feature originates from a real or translated modality. Through adversarial optimization, $E_{\mathrm{I}}$ is encouraged to make the two domains indistinguishable in feature space.

\begin{figure}
    \centering
    \includegraphics[width=\linewidth]{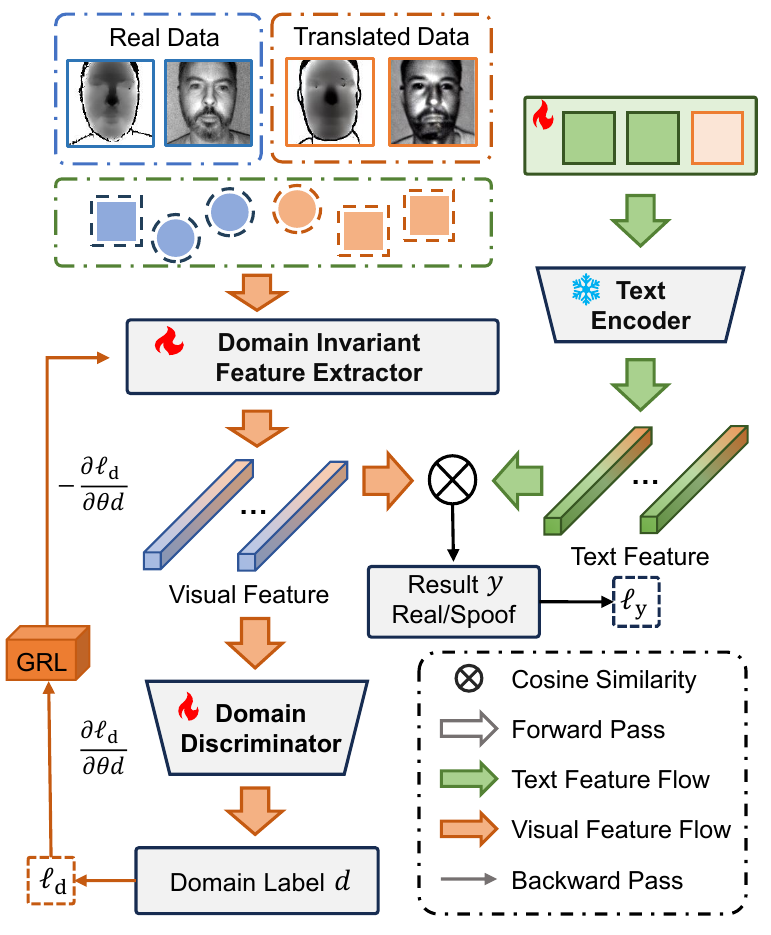 }
    \caption{Illustration of the proposed Synthetic-to-Real Face Domain Alignment (SR-FDA).
    By leveraging adversarial learning between real and synthetic data domains, SR-FDA extracts domain-invariant features to alleviate domain shift introduced by generated data, while aligning visual and textual features for robust real vs. spoof classification.
     The source face images are from the WMCA dataset~\cite{george2019biometric}.}
    \label{fig:domain_generalization}
\end{figure}

As illustrated in Fig.\ref{fig:domain_generalization}, SR-FDA consists of the image feature extractor $E_{\mathrm{I}}$, the text feature extractor $E_{\mathrm{T}}$, the classifier $\omega$, and the domain discriminator $d$. Different from generic domain-adversarial adaptation, SR-FDA defines the adversarial domains by modality origin rather than by subject identity or dataset source. Therefore, the discriminator is used specifically to reduce the synthetic-to-real discrepancy introduced by FMIT, while the fPAD classifier remains optimized for bona fide/spoof separation.

    \begin{equation}
    \begin{aligned}
         \hat{y} = \omega \left( F, \{\hat{w}_i\}_{i=1}^{K}\right)
    \end{aligned}
    \end{equation}

The classifier uses visual--textual similarity. Classification gradients update the learnable detector prompts and the trainable visual parameters, including F\textsuperscript{2}A; they do not update the separately trained FMIT. The domain branch operates on visual features, so its reversed gradients constrain the trainable visual parameters rather than directly updating the text prompts.

The classification task adopts the Cross-Entropy Loss to measure the discrepancy between the predicted label distribution $q(x_i)$ and the ground-truth label distribution $p(x_i)$. The classification loss $\ell_{\mathrm{y}}$ is computed as follows:

    \begin{equation}
    \ell_y = - \sum_{i = 1}^{n} p(x_i) \log \left( q(x_i) \right)
    \end{equation}

Due to the domain shift between translated auxiliary observations and real observations, a domain discriminator $d$ is introduced for modality-origin classification. The objective of $d$ is to determine whether the input feature originates from the real-observation domain $D_{s}^{r}$ or the translated-observation domain $D_{s}^{g}$. Specifically, the domain discriminator $d$ distinguishes between image features $F_r$ from real observations and $F_g$ from translated observations. The domain discrimination objective $\ell_d$ is defined as follows:

    \begin{equation}
    \ell_{\text{d}}(X_r, X_g) = \mathbb{E} \left[ \log \left( d(F_r) \right) \right] + \mathbb{E} \left[ \log \left( 1 - d(F_g) \right) \right]
    \end{equation}

A Gradient Reversal Layer (GRL)\cite{ganin2016domain} is introduced between the domain discriminator and the image feature extractor. During the forward pass, GRL acts as an identity mapping; during backpropagation, it multiplies the gradient from the domain discriminator by a negative coefficient. This mechanism enables joint adversarial training: the discriminator learns to identify modality origin, whereas the image encoder is optimized to remove modality-origin information from the visual representation.

Here, $\ell_d$ is a domain log-likelihood, which the discriminator maximizes, equivalently minimizing the negative log-likelihood. The reversed gradient encourages the trainable visual parameters to reduce this log-likelihood while also minimizing the classification loss. Only trainable parameters receive updates; the notation below describes the branch objectives, not separate training stages. The classification-side parameters include the learnable prompts used to construct the text representations.

The classification branch minimizes the fPAD classification loss to optimize the parameters $\theta_{E_{\mathrm{I}}}$ and $\theta_{\omega}$, which can be expressed as:

    \begin{equation}
    \hat{\theta}_{E_{\mathrm{I}}}, \hat{\theta}_{\omega} = \arg\min_{\theta_{E_{\mathrm{I}}}, \, \theta_{\omega}} \, \ell_y \left( \theta_{E_{\mathrm{I}}}, \theta_{\omega} \right)
    \end{equation}

Meanwhile, the domain branch optimizes the domain discriminator parameters $\theta_d$ to separate real and translated observations, while the GRL reverses this gradient before it reaches $E_{\mathrm{I}}$. The domain objective can be formulated as:

    \begin{equation}
    \hat{\theta}_{\text{d}} = \arg\max_{\theta_{\text{d}}} \, \ell_{\text{d}} \left( \hat{\theta}_{E_{\mathrm{I}}}, \theta_{\text{d}} \right)
    \end{equation}

\section{Experiments}

\subsection{Experiment Setup}

\textbf{Implementation details.}
All experiments were conducted using the PyTorch framework on a GPU with 24\,GB of memory. FMIT was initialized from Stable Diffusion v1.5 and trained for conditional modality translation using RGB observations and target-modality instructions. The denoising UNet learns to predict the noise added to the target-modality latent representation, conditioned on the RGB observation and instruction. The variational autoencoder (VAE) and text encoder were frozen, and the UNet was optimized using AdamW with a learning rate of $5\times10^{-5}$. Input images were resized to $256\times256$, and mixed-precision training, gradient checkpointing, random horizontal flipping, and a conditioning dropout probability of 0.05 were adopted.

FMIT is trained separately and remains frozen while the detector is trained. The learnable detector prompts, F\textsuperscript{2}A, and the SR-FDA domain discriminator are optimized jointly during detector training. The detector was trained using Adam with a batch size of 16. For the fPAD detector, we used CLIP ViT-B/16 as the visual backbone with an input resolution of $224\times224$. The prompt length was set to 16, class-specific context learning was enabled, and the class token was placed in the middle of the learnable context tokens.

The F\textsuperscript{2}A module was implemented as a lightweight adapter inserted into the visual Transformer blocks. It uses a bottleneck projection and a $3\times3$ convolution on the reshaped patch-token grid to capture local spoof-relevant cues with limited additional parameters. SR-FDA was implemented with a lightweight domain discriminator and auxiliary domain-classification losses for depth and IR features, jointly optimized with the fPAD classification loss.

\textbf{Datasets.} We employed the WMCA and PADISI-Face datasets to evaluate the proposed method under multimodal and flexible-modality protocols. The WMCA dataset~~\cite{george2019biometric} consists of 1,679 videos from 72 subjects, recorded in four modalities: RGB, depth, IR, and thermal. It covers diverse presentation attack types including print, replay, glasses, fake heads, hard masks, flexible masks, and paper masks. Data were captured using an Intel RealSense SR300 (RGB, NIR, depth) and a Thermal Compact PRO (thermal).

The PADISI-Face dataset~~\cite{rostami2021detection} contains 2,029 images from 182 participants, collected with a multi-sensor array spanning RGB, short-wave infrared (SWIR), and long-wave infrared (thermal). It includes a wide range of attack types such as disguise, print, 3D masks, and replay. The acquisition setup employed synchronized LED illumination across different wavelengths to ensure temporal consistency and maximize sensor throughput.
Experiments were conducted under both intra-domain and cross-domain settings across the two datasets to assess robustness and generalization under varying modality availability and domain shifts.

This study uses the existing WMCA and PADISI-Face datasets for secondary analysis. No participants were recruited, and no new human-subject data were collected for this study.

\subsection{Evaluation Protocol}
To comprehensively evaluate model performance under various modality settings, we design a flexible-modality evaluation protocol based on ~\cite{yu2023flexible} and ~\cite{yu2024visual}, covering both intra-domain and cross-domain evaluations, using the WMCA and PADISI-Face datasets.

\textbf{Intra-domain evaluation.} The model is trained and tested on the same dataset. The fixed-multimodal task uses complete RGB, IR, and depth observations. In the flexible-modality task, missing auxiliary modalities are simulated during training and testing according to P1--P4 below.

\textbf{Cross-domain evaluation.} We evaluate both WMCA-to-PADISI-Face and PADISI-Face-to-WMCA transfer under sub-modality settings. In each direction, FMIT and the detector are trained using only the source-domain training data; no target-domain data are used for training. The trained FMIT is kept frozen when translating target-domain inputs at evaluation time. For the flexible-modality task in both intra-domain and cross-domain evaluation, we introduce random modality missingness during training and testing. RGB is always retained. For each auxiliary modality subject to missingness, $\alpha$ denotes its missing probability, set to 70\% or 100\%.

\begin{figure}
     \centering
     \includegraphics[width=\linewidth]{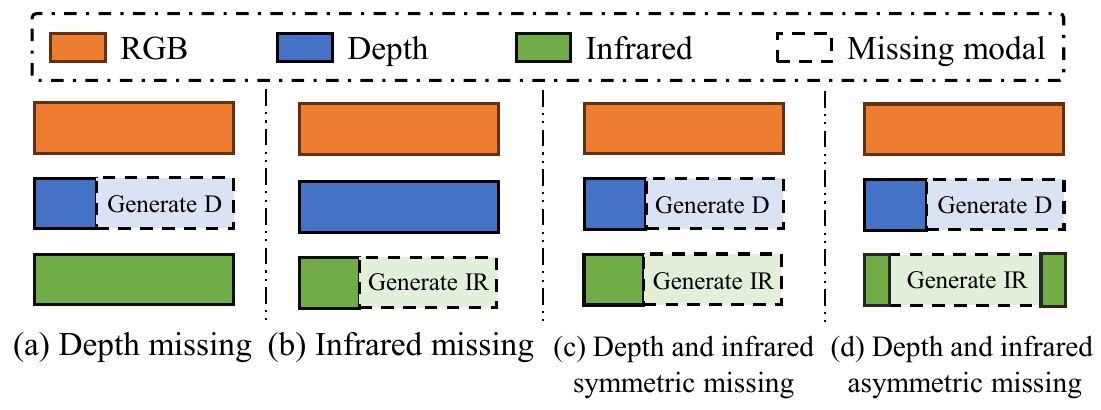}
     \caption{Illustration of flexible-modality translation and completion protocols: (a) depth-only missing, (b) IR-only missing, (c) symmetric missing of both depth and IR modalities, and (d) asymmetric missing of depth and IR modalities.}
     \label{fig:flexible_protocol}
 \end{figure}

Four modality-missing protocols are defined (Fig.~\ref{fig:flexible_protocol}, panels (a--d) correspond to P1--P4):
P1: RGB + IR with partial depth missing;
P2: RGB + depth with partial IR missing;
P3: RGB with a shared missingness mask for IR and depth, so that both are missing together with probability $\alpha$;
and P4: RGB with independently sampled missingness masks for IR and depth, each with missing probability $\alpha$.
When the missing ratio $\alpha=100\%$, P3 and P4 become equivalent and are jointly denoted as P3\&P4.

To address missing modalities, we adopt a flexible-modality translation strategy, dynamically generating the absent modalities based on sample conditions. The completed multimodal inputs are then fed into the fPAD model for training and inference, enhancing robustness and generalization in incomplete modality scenarios.

\subsection{Fixed Modality Scenario}
As shown in Table \ref{tab:unseen_wmca}, we evaluate the proposed method on WMCA under both the visible-protocol (Grandtest) and the unseen-protocol (Leave-One-Out, LOO) settings, and compare it with eight state-of-the-art approaches to assess detection generalization to unknown presentation-attack types. Bold values indicate the best performance, and underlined values denote the second-best. In the LOO protocol, attack IDs A1--A7 correspond to flexible mask, replay, fake head, print, glasses, paper mask, and hard mask attacks, respectively. FMA-Net achieves the lowest mean ACER across the seven unseen-attack protocols (1.21\%), compared with 2.17\% for CPL-CLIP. Its sample standard deviation across these protocols is 2.03 percentage points. Notably, the ACER value for visible attacks is only 0.66\%, while for the unseen flexible mask attack protocol, the ACER value is 0.37\%. The ACER values for unseen head mask and print attacks are both 0, with unseen paper mask attacks yielding an ACER of 0.24\% and unseen hard mask attacks yielding the best result.

The relatively weaker result on unseen replay attacks suggests that replay-specific artifacts remain challenging under the proposed representation. Unlike mask or print attacks, replay attacks often exhibit subtle display-related cues, such as screen reflection, moire patterns, temporal blur, and boundary artifacts, which may be weakly expressed in still multimodal observations. Since FMA-Net focuses on translated auxiliary observations and local--global facial representation adaptation, these fine-grained display artifacts may be partially smoothed by modality translation or less emphasized by the learned adapter. This observation indicates a limitation of the current framework for attacks whose discriminative evidence is dominated by high-frequency reflection and display cues.

\begin{table*}[!t]
\caption{Comparative experimental results of different attack protocols on the WMCA dataset (ACER,\%)}
\label{tab:unseen_wmca}
\centering
\resizebox{\textwidth}{!}{
\begin{tabular}{l c c c c c c c c c}
\toprule
\multirow{2}{*}[-0.6ex]{Method} & \multirow{2}{*}[-0.6ex]{Seen} & \multicolumn{7}{c}{Unseen } & \multirow{2}{*}[-0.6ex]{\textbf{Mean$\pm$Std}}\\
\cmidrule{3-9}
& & A1 & A2 & A3 & A4 & A5 & A6 & A7 & \\
\midrule
MC-PixBiS~\cite{george2019deep} & 1.80 & 49.70 & 3.70 & 0.70 & \underline{0.10} & 16.00 & \underline{0.20} & 3.40 & 10.54$\pm$18.14 \\
MCCNN-OCCL-GMM~\cite{george2020learning} & 3.30 & 22.80 & 31.40 & 1.90 & 30.00 & 50.00 & 4.80 & 18.30 & 22.74$\pm$16.57 \\
CMFL~\cite{george2021cross} & 1.70 & 12.40 & 1.00 & 2.50 & 0.70 & 33.50 & 1.80 & 1.70 & 7.66$\pm$12.11 \\
ViT~\cite{liu2024fm} & 2.71 & 11.95 & 1.44 & 3.78 & \textbf{0.00} & 18.02 & 0.58 & 4.43 & 5.74$\pm$6.75 \\
MA-ViT~\cite{xiao2024} & 1.45 & 9.76 & 0.93 & \underline{0.55} & \textbf{0.00} & 14.00 & \textbf{0.00} & 1.46 & 3.81$\pm$5.67 \\
FM-ViT~\cite{liu2023fm} & 1.00 & 3.56 & 0.72 & \textbf{0.00} & \textbf{0.00} & 12.00 & 0.43 & 0.73 & 2.49$\pm$4.37 \\
FM-CLIP~\cite{liu2024fm} & 1.05 & 3.35 & \underline{0.69} & \textbf{0.00} & \textbf{0.00} & \underline{11.00} & 0.38 & 0.66 & 2.30$\pm$4.01 \\
CPL-CLIP~\cite{zhang2024cpl} & \underline{0.79} & \underline{1.37} & \textbf{0.00} & \textbf{0.00} & \textbf{0.00} & 13.28 & \textbf{0.00} & \underline{0.53} & \underline{2.17}$\pm$4.93 \\
FMA-Net(ours) & \textbf{0.66} & \textbf{0.37} & 2.32 & \textbf{0.00} & \textbf{0.00} & \textbf{5.42} & 0.24 & \textbf{0.14} & \textbf{1.21}$\pm$2.03 \\
\bottomrule
\end{tabular}
}
\par\smallskip
{\footnotesize\raggedright Seen: Grandtest. A1--A7 (held-out attacks): flexible mask, replay, fake head, print, glasses, paper mask, and hard mask. Mean$\pm$Std summarizes Unseen ACERs only (sample standard deviation).\par}
\end{table*}

To further validate the performance of the proposed FMA-Net method when facing unseen attacks, an unseen attack protocol similar to the WMCA dataset was established on the PADISI-Face dataset. The experimental results are shown in Table \ref{tab:unseen_padis}. The held-out attack categories A1--A9 represent disguised eye attacks, paper attacks, fake mask attacks, half-mask attacks, transparent mask attacks, makeup attacks, partial paper glasses attacks, silicone mask attacks, and tattoo attacks. The input modalities include RGB, depth, and IR. In the comparison methods, "-H" and "-E" represent the Hyp and Euc methods from the literature ~\cite{han2023hyperbolic}, respectively.

As shown in Table \ref{tab:unseen_padis}, FMA-Net achieves a mean ACER of 3.17\% across the nine unseen-attack protocols, the lowest among the listed RDI methods and the second-lowest across all listed configurations. MC-PixBiS-E-H with RD inputs obtains the lowest mean ACER of 3.15\%. FMA-Net achieves an ACER of 0.00\% for fake mask, transparent mask, and partial paper glasses attacks, and a tied second-lowest ACER of 0.14\% for seen attacks.

FMA-Net performs well in most mask-related attacks, indicating its effectiveness in detecting attacks with abnormal facial boundaries and geometric discontinuities. However, eyewear-style attacks remain more challenging because their spoof traces are spatially limited and can be visually similar to bona fide glasses. In such cases, the decision relies on subtle cues around the lens boundary, reflection pattern, and local material response, rather than on large-scale facial deformation. These localized cues may contribute to the variation in performance across attack types. The result suggests that the proposed modality translation and feature adaptation strategy is effective for broad facial spoof artifacts, while future improvements should further strengthen localized material-aware modeling for eyewear-like attacks.

\begin{table*}[!t]
\caption{Comparative experimental results of different attack protocols on the PADISI-Face dataset (ACER,\%)}
\label{tab:unseen_padis}
\centering
\resizebox{\textwidth}{!}{
\begin{tabular}{l c c c c c c c c c c c c}
\toprule
\multirow{2}{*}[-0.6ex]{\textbf{Modality}} & \multirow{2}{*}[-0.6ex]{\textbf{Method}} & \multirow{2}{*}[-0.6ex]{\textbf{Seen}} & \multicolumn{9}{c}{\textbf{Unseen}} & \multirow{2}{*}[-0.6ex]{\textbf{Mean$\pm$Std}} \\
\cmidrule{4-12}
& & & A1 & A2 & A3 & A4 & A5 & A6 & A7 & A8 & A9 & \\
\midrule
\multirow{3}{*}{R} & ViT~\cite{liao2023domain} & 0.28 & 22.33 & 0.27 & 0.54 & 0.54 & 0.27 & 9.49 & 41.18 & 3.99 & 1.80 & 8.93$\pm$14.08 \\
& ViT-H~\cite{han2023hyperbolic} & \underline{0.14} & 24.66 & 0.27 & 0.41 & \underline{0.27} & 0.27 & 13.12 & 32.36 & 3.04 & 1.67 & 8.45$\pm$12.23 \\
& ViT-H-H~\cite{han2023hyperbolic} & \underline{0.14} & 18.04 & 0.27 & 0.54 & \textbf{0.14} & \underline{0.14} & 9.91 & 10.17 & 5.14 & 2.07 & 5.16$\pm$6.31 \\
\cmidrule{1-13}
\multirow{4}{*}{RD} & ViT-E-E~\cite{han2023hyperbolic} & 0.55 & 24.43 & 0.27 & \underline{0.27} & 0.41 & \underline{0.14} & 16.2 & 14.31 & 3.18 & \underline{0.41} & 6.62$\pm$9.22 \\
& ViT-E-H~\cite{han2023hyperbolic} & \underline{0.14} & 20.54 & \textbf{0.00} & 0.41 & 0.41 & \underline{0.14} & 8.37 & 9.50 & 5.14 & \textbf{0.27} & 4.98$\pm$6.95 \\
& MC-PixBiS-E-E~\cite{han2023hyperbolic} & 0.33 & 4.53 & 0.54 & 0.54 & \textbf{0.14} & 0.41 & 5.43 & 32.63 & \textbf{0.41} & 0.95 & 5.06$\pm$10.52 \\
& MC-PixBiS-E-H~\cite{han2023hyperbolic} & \underline{0.14} & \textbf{1.35} & \underline{0.14} & \underline{0.27} & 0.95 & 0.27 & 5.70 & 18.45 & \textbf{0.41} & 0.81 & \textbf{3.15}$\pm$5.99 \\
\cmidrule{1-13}
\multirow{4}{*}{RI} & ViT-E-E~\cite{han2023hyperbolic} & 0.30 & 22.74 & \textbf{0.00} & 0.41 & \underline{0.27} & 0.27 & 13.12 & 23.27 & 2.91 & \textbf{0.27} & 7.03$\pm$9.97 \\
& ViT-E-H~\cite{han2023hyperbolic} & 0.28 & 22.23 & \textbf{0.00} & 0.41 & \underline{0.27} & \textbf{0.00} & 11.17 & 5.49 & 3.31 & 0.68 & 4.84$\pm$7.50 \\
& MC-PixBiS-E-E~\cite{han2023hyperbolic} & 0.28 & 3.72 & 0.41 & 0.54 & 0.54 & 0.64 & \underline{3.75} & 33.44 & \underline{0.54} & 2.07 & 5.07$\pm$10.73 \\
& MC-PixBiS-E-H~\cite{han2023hyperbolic} & \textbf{0.00} & \underline{3.01} & \underline{0.14} & \underline{0.27} & 0.68 & 0.41 & 6.96 & 23.13 & \textbf{0.41} & 2.21 & 4.14$\pm$7.45 \\
\cmidrule{1-13}
\multirow{3}{*}{RDI} & CoCoOp~\cite{zhou2022conditional} & 0.47 & 21.40 & 15.25 & 1.68 & 0.96 & 1.24 & 6.76 & 0.28 & 5.10 & 0.68 & 5.93$\pm$7.52 \\
& CoOp~\cite{zhou2022learning} & 0.71 & 13.17 & 0.48 & 0.70 & 0.95 & 7.63 & \textbf{2.11} & \underline{0.17} & 12.32 & 0.79 & 4.26$\pm$5.33 \\
& FMA-Net(ours) & \underline{0.14} & 12.99 & 0.28 & \textbf{0.00} & \textbf{0.14} & \textbf{0.00} & 7.04 & \textbf{0.00} & 7.58 & 0.54 & \underline{3.17}$\pm$4.81 \\
\bottomrule
\end{tabular}
}
\par\smallskip
{\footnotesize\raggedright A1--A9 (held-out attacks): disguised eye, paper, fake mask, half-mask, transparent mask, makeup, partial paper glasses, silicone mask, and tattoo. R/D/I: RGB/depth/infrared. Mean$\pm$Std summarizes Unseen ACERs only (sample standard deviation).\par}
\end{table*}

\subsection{Flexible Modality Scenario}

\paragraph{Intra-Domain Test}

Experiments were conducted on the WMCA dataset with missing modality ratios of 70\% and 100\%, and the results are shown in Table \ref{tab:flexible_wmca_70_100}. As can be seen, under the 70\% setting, the proposed FMA-Net method achieves the best ACER values across all four protocols when compared to five commonly used existing methods. In particular, for single missing modality scenarios, FMA-Net obtains ACER values of 2.15\% and 2.78\% in Protocols 1 and 2, significantly outperforming the second-best VP-FAS method, especially in Protocol 1 where the improvement reaches 1.52 percentage points (3.67\% vs. 2.15\%). Averaged over P1--P4, FMA-Net reduces ACER from 4.24\% for VP-FAS to 3.22\%. For two missing modalities, FMA-Net also achieves the best results, though the margin is relatively smaller. This demonstrates that face modality completion using translation can effectively enhance presentation attack detection.

When the missing modality ratio is increased to 100\%, FMA-Net is compared against three flexible-modality methods (ViT ~\cite{george2020learning}, CoCoOp ~\cite{zhou2022conditional}, CoOp ~\cite{zhou2022learning}) and seven traditional multimodal methods (MA-Net ~\cite{liu2021face}, MC-PixBiS ~\cite{george2019deep}, CMFL ~\cite{george2021cross}, MA-ViT ~\cite{xiao2024}, CCL(Resnet) ~\cite{liu2022contrastive}, CCL(CDCN) ~\cite{liu2022contrastive}, MC-CNN ~\cite{george2020learning}). The proposed method achieves the best ACER values in Protocols 3\&4, reducing the ACER by 0.76 percentage points compared to the second-best result (5.23\% vs. 5.99\%). In Protocol 1, FMA-Net attains second-best performance, while in Protocol 2 it ties with CMFL for second-best performance, with ACER values slightly higher than the best results by 0.40 percentage points (2.55\% vs. 2.15\%) and 0.25 percentage points (1.70\% vs. 1.45\%), respectively. Notably, in Protocol 2, FMA-Net outperforms all flexible-modality methods but slightly lags behind the traditional multimodal MA-ViT, reflecting the inherent challenges of flexible-modality translation.

By comparing the two settings, it is observed that in Table \ref{tab:flexible_wmca_70_100}, the average ACER for Protocols 3\&4 is higher than that of Protocols 1\&2, with an increase of 1.50 percentage points (3.97\% vs. 2.47\%) under the 70\% setting, and 3.10 percentage points (5.23\% vs. 2.13\%) under the 100\% setting. This trend is expected because Protocols 3\&4 rely more heavily on translated auxiliary observations, whereas Protocols 1\&2 still preserve one non-RGB modality. In particular, Protocol 2 retains depth but lacks IR, which can be less favorable for attacks whose evidence is strongly associated with spectral reflection or material response. Therefore, a moderate amount of modality translation can provide useful complementary cues, while a high missing ratio increases the dependence on translated observations and may amplify translation uncertainty.

\begin{table}[htbp!]
\caption{Comparison results for the flexible-modality translation completion protocol on the WMCA dataset (ACER,\%)}
\label{tab:flexible_wmca_70_100}
\centering
\resizebox{\columnwidth}{!}{
\begin{tabular}{lcccc}
\toprule
Method & P1 (RI) & P2 (RD) & P3 (R) & P4 (R) \\
\midrule
\multicolumn{5}{c}{\bfseries $\alpha=70\%$} \\
\midrule
ViT~\cite{george2020learning}           & 10.61 & 5.86 & 8.70 & 6.37 \\
Prompt~\cite{jia2022visual}         &  8.70 & 5.52 & 5.84 & 7.21 \\
CMFL~\cite{george2021cross}           &  6.53 & 4.95 & 5.08 & 5.37 \\
CrossAtten~\cite{yu2023flexible}     &  8.47 & 5.29 & 8.33 & 7.15 \\
VP-FAS~\cite{yu2024visual}         & \underline{3.67} & \underline{3.91} & \underline{4.54} & \underline{4.83} \\
FMA-Net(ours)                               & \textbf{2.15} & \textbf{2.78} & \textbf{3.88} & \textbf{4.05} \\
\midrule
\multicolumn{5}{c}{\bfseries $\alpha=100\%$} \\
\midrule
MA-Net~\cite{liu2021face}        & 6.85 & -    & \multicolumn{2}{c}{-} \\
MC-PixBiS~\cite{george2019deep}  & -    & 1.80 & \multicolumn{2}{c}{-} \\
CMFL~\cite{george2021cross}      & -    & \underline{1.70} & \multicolumn{2}{c}{-} \\
MA-ViT~\cite{xiao2024}           & -    & \textbf{1.45} & \multicolumn{2}{c}{-} \\
CCL(Resnet)~\cite{liu2022contrastive} & - & - & \multicolumn{2}{c}{30.69} \\
CCL(CDCN)~\cite{liu2022contrastive}   & - & - & \multicolumn{2}{c}{27.14} \\
MC-CNN~\cite{george2020learning} & -    & -    & \multicolumn{2}{c}{32.82} \\
ViT~\cite{george2020learning}    & 4.02 & 3.10 & \multicolumn{2}{c}{9.84} \\
CoCoOp~\cite{zhou2022conditional}& 3.46 & 3.76 & \multicolumn{2}{c}{7.60} \\
CoOp~\cite{zhou2022learning}     & \textbf{2.15} & 2.65 & \multicolumn{2}{c}{\underline{5.99}} \\
FMA-Net(ours)                   & \underline{2.55} & \underline{1.70} & \multicolumn{2}{c}{\textbf{5.23}} \\
\bottomrule
\end{tabular}
}
\par\smallskip
{\footnotesize\raggedright P1: depth missing; P2: IR missing; P3/P4: shared/independent depth--IR missingness.\par}
\end{table}

Experiments were conducted on the PADISI-Face dataset with modality missing ratios of 70\% and 100\% for flexible-modality translation completion, and the results are shown in Table \ref{tab:flexible_padisi_70_100}. As observed, the proposed FMA-Net method performs best in Protocols 1 and 2 under the 70\% setting, with ACER values of 0.48\% and 0.74\%, respectively. It achieves the second-best result in Protocol 3 (0.74\% vs. 0.69\%). In Protocol 4, however, its ACER of 1.43\% is higher than those of CoCoOp (0.90\%) and CoOp (1.24\%) by 0.53 and 0.19 percentage points, respectively. When the missing ratio is increased to 100\%, the proposed method achieves the best ACER values in Protocols 1 and 3\&4, and the second-best result in Protocol 2 with only a 0.20 percentage point gap (1.11\% vs. 0.91\%). These results indicate that translated modalities provide useful complementary information, but the benefit depends on the missing pattern and attack composition. The relatively modest margins in some protocols suggest that when the available evidence is already sufficient or when the missing modality contains highly localized cues, translated observations may introduce uncertainty into feature extraction. This is a possible explanation for the protocol-dependent performance of FMA-Net.

\begin{table}[!t]
\caption{Comparison results for the flexible-modality translation completion protocol on the PADISI-Face dataset (ACER,\%)}
\label{tab:flexible_padisi_70_100}
\centering
\resizebox{\columnwidth}{!}{
\begin{tabular}{lcccc}
\toprule
Method & P1 (RI) & P2 (RD) & P3 (R) & P4 (R) \\
\midrule
\multicolumn{5}{c}{\bfseries $\alpha=70\%$} \\
\midrule
CoCoOp~\cite{zhou2022conditional} & 0.80 & 0.91 & \textbf{0.69} & \textbf{0.90} \\
CoOp~\cite{zhou2022learning}      & 0.55 & 1.36 & 0.85 & 1.24 \\
FMA-Net(ours)                    & \textbf{0.48} & \textbf{0.74} & 0.74 & 1.43 \\
\midrule
\multicolumn{5}{c}{\bfseries $\alpha=100\%$} \\
\midrule
CoCoOp~\cite{zhou2022conditional} & 0.44 & 2.20 & \multicolumn{2}{c}{0.93} \\
CoOp~\cite{zhou2022learning}      & 0.99 & \textbf{0.91} & \multicolumn{2}{c}{1.07} \\
FMA-Net(ours)                    & \textbf{0.39} & 1.11 & \multicolumn{2}{c}{\textbf{0.91}} \\
\bottomrule
\end{tabular}
}
\par\smallskip
{\footnotesize\raggedright Protocols follow Table~\ref{tab:flexible_wmca_70_100}.\par}
\end{table}

\paragraph{Cross-Domain Test}

Tables \ref{tab:flexible_outra_70} and \ref{tab:flexible_outra_100} compare cross-domain detection with and without SR-FDA. At a 70\% missing ratio, alignment improves W-P P1 from 9.95\% to 8.99\%, but worsens the other seven reported settings, including W-P P3 from 9.72\% to 22.01\%. At a 100\% missing ratio, it improves P-W P1 and P3\&P4, but worsens the other four settings. Thus, its cross-domain benefit is limited and protocol-dependent. SR-FDA aligns real and translated observations within the training distribution; this objective is distinct from alignment between datasets. One possible explanation is that FMIT retains source-domain appearance or sensor characteristics in the translated observations, introducing a mismatch on the target dataset. Alignment learned on these source-domain observations may not transfer to the target domain and may suppress features useful for target-domain attack discrimination. These are possible explanations, not mechanisms established by the reported performance comparisons.

\begin{table}[!t]
\caption{The cross-domain comparison results for flexible-modality translation completion with \(\alpha = 70\%\) (ACER,\%)}
\label{tab:flexible_outra_70}
\centering
\resizebox{\columnwidth}{!}{
\begin{tabular}{l c c c c c c c c}
\toprule
\multirow{2}{*}{Method}  & \multicolumn{4}{c}{W-P} & \multicolumn{4}{c}{P-W} \\
\cmidrule{2-5} \cmidrule{6-9}
& P1 & P2&P3&P4 & P1 & P2&P3&P4 \\
\midrule
CoCoOp~\cite{zhou2022conditional} & 10.99 & 28.75 & 9.86 & 16.07 & \textbf{18.93} & 22.11 & 18.12 & 17.25 \\
CoOp~\cite{zhou2022learning} & 14.25 & 20.30 & 11.29 & 12.95 & 19.47 & 18.56 & 19.25 & 19.36 \\
FMA-Net w/o SR-FDA (ours) & 9.95 & \textbf{19.80} & \textbf{9.72} & \textbf{11.29} & 19.46 & \textbf{15.79} & \textbf{17.89} & \textbf{16.63}\\
FMA-Net w/ SR-FDA (ours) & \textbf{8.99} & 21.84 & 22.01 & 20.93 & 19.80 & 19.75 & 18.87 & 17.94 \\
\bottomrule
\end{tabular}
}
\par\smallskip
{\footnotesize\raggedright Protocols follow Table~\ref{tab:flexible_wmca_70_100}. W-P: WMCA to PADISI-Face; P-W: reverse direction.\par}
\end{table}

The cross-domain experimental results for flexible-modality translation completion with a modality missing ratio of 100\% are shown in Table \ref{tab:flexible_outra_100}. In addition to comparisons with flexible-modality methods, the results are also compared with the cross-domain results of four multimodal methods. As observed, FMA-Net without SR-FDA maintains competitive performance in several W-P protocols, while FMA-Net with SR-FDA obtains the best result in Protocols 3\&4 of the P-W cross-domain task, with an advantage of 0.29 percentage points (16.99\% vs. 17.28\%). Nevertheless, the overall cross-domain advantage becomes less consistent when all non-RGB modalities are unavailable. In P3\&P4 at 100\% missingness, the model relies entirely on RGB and translated auxiliary observations; P1 and P2 retain one acquired non-RGB modality. Under such conditions, translated modalities can still provide useful structural and spectral priors, but they cannot fully replace physically acquired modalities, especially when the target domain exhibits different sensor characteristics or attack materials. These results reveal an important limitation of the current framework: modality translation is effective as a compensation mechanism, but its reliability decreases when both the missing ratio and domain gap are large.

\begin{table}[!t]
\caption{The cross-domain comparison results for flexible-modality translation completion with \(\alpha = 100\%\) (ACER,\%)}
\label{tab:flexible_outra_100}
\centering
\resizebox{\columnwidth}{!}{
\begin{tabular}{l c c c c c c}
\toprule
\multirow{2}{*}{Method}  & \multicolumn{3}{c}{W-P} & \multicolumn{3}{c}{P-W} \\
\cmidrule{2-4} \cmidrule{5-7}
& P1 & P2 & P3\&P4 & P1 & P2 & P3\&P4 \\
\midrule
ViT-E-E~\cite{han2023hyperbolic} & 11.78 & 11.90 & - & \underline{13.14} & 15.39  & - \\
ViT-E-H~\cite{han2023hyperbolic} & 11.29 & \textbf{8.50} & -  & \textbf{12.55} & \textbf{12.05} & - \\
MC-PixBiS-E-E~\cite{han2023hyperbolic} & 20.06 & 16.68  & -  & 15.90 & 15.51& - \\
MC-PixBiS-E-H~\cite{han2023hyperbolic} & 13.82 & 15.37 & -  & 15.63 & \underline{14.20} & - \\
CoCoOp~\cite{zhou2022conditional} & \underline{10.87} & 28.75 & 16.07 & 18.93 & 19.12 & 17.89 \\
CoOp~\cite{zhou2022learning} & 12.11 & 12.44 & \underline{13.52} & 22.11 & 18.03 & \underline{17.28} \\
FMA-Net w/o SR-FDA (ours) & \textbf{9.20} & \underline{10.55} & \textbf{10.61} & 18.82 & 17.76 & 18.82 \\
FMA-Net w/ SR-FDA (ours) & 11.30 & 20.67 & 21.75 & 18.55 & 18.70 & \textbf{16.99} \\
\bottomrule
\end{tabular}
}
\par\smallskip
{\footnotesize\raggedright Protocols follow Table~\ref{tab:flexible_wmca_70_100}. Transfer directions follow Table~\ref{tab:flexible_outra_70}.\par}
\end{table}

\subsection{Ablation Study}

When the modality missing ratio is set to 70\%, the experimental results in Table \ref{tab:ablation_flexible_70} show that incorporating SR-FDA in the flexible-modality translation completion protocol outperforms the baseline network and the variant without domain alignment. Adding F\textsuperscript{2}A reduces ACER in Protocols 1, 3, and 4 by 1.07, 1.33, and 1.47 percentage points, respectively. In Protocol 2, however, ACER increases from 2.92\% to 3.16\%, indicating that the benefit of feature adaptation is protocol-dependent. Notably, after adding SR-FDA, the ACER values are further reduced across all protocols, with the largest reduction of 0.71 percentage points (2.15\% vs. 2.86\%). These results support the benefit of SR-FDA in the evaluated intra-domain setting, although the ACER comparisons alone do not establish the mechanism of improvement.

\begin{table}[!t]
\caption{Ablation study of Prompt Learning, F\textsuperscript{2}A, and SR-FDA under the flexible-modality translation completion protocol with \(\alpha = 70\%\) (ACER,\%)}
\label{tab:ablation_flexible_70}
\centering
\resizebox{\columnwidth}{!}{
\begin{tabular}{c c c c c c c}
\toprule
Prompt & F\textsuperscript{2}A & SR-FDA & P1 (RI) & P2 (RD) & P3 (R) & P4 (R) \\
\midrule
\checkmark &  &  & 3.93 & 2.92 & 5.79 & 5.88 \\
\checkmark & \checkmark &  & 2.86 & 3.16 & 4.46 & 4.41 \\
\checkmark & \checkmark & \checkmark & \textbf{2.15} & \textbf{2.78} & \textbf{3.88} & \textbf{4.05} \\
\bottomrule
\end{tabular}
}
\par\smallskip
{\footnotesize\raggedright Protocols follow Table~\ref{tab:flexible_wmca_70_100}. Results are on WMCA; all variants use FMIT completion.\par}
\end{table}

With a modality missing ratio of 100\%, Table \ref{tab:ablation_flexible_100} shows that adding SR-FDA reduces ACER from 2.03\% to 1.70\% in P2 and from 5.87\% to 5.23\% in P3\&P4. In P1, however, ACER increases from 2.11\% to 2.55\%, a deterioration of 0.44 percentage points. Thus, alignment improves the settings where IR alone or both auxiliary modalities are missing, but is detrimental in the depth-missing setting evaluated here. These results show that the benefit of SR-FDA depends on the missing-modality pattern; they do not establish the mechanism responsible for the P1 degradation.

\begin{table}[!t]
\caption{Ablation study of Prompt Learning, F\textsuperscript{2}A, and SR-FDA under the flexible-modality translation completion protocol with \(\alpha = 100\%\) (ACER,\%)}
\label{tab:ablation_flexible_100}
\centering
\resizebox{\columnwidth}{!}{
\begin{tabular}{c c c c c c}
\toprule
Prompt & F\textsuperscript{2}A & SR-FDA & P1 (RI) & P2 (RD) & P3\&P4 (R) \\
\midrule
\checkmark &  &  & 2.15 & 2.65 & 5.99 \\
\checkmark & \checkmark &  & \textbf{2.11} & 2.03 & 5.87 \\
\checkmark & \checkmark & \checkmark & 2.55 & \textbf{1.70} & \textbf{5.23} \\
\bottomrule
\end{tabular}
}
\par\smallskip
{\footnotesize\raggedright Dataset, protocols, and FMIT completion follow Table~\ref{tab:ablation_flexible_70}.\par}
\end{table}

\begin{figure}
    \centering
    \includegraphics[width=\linewidth]{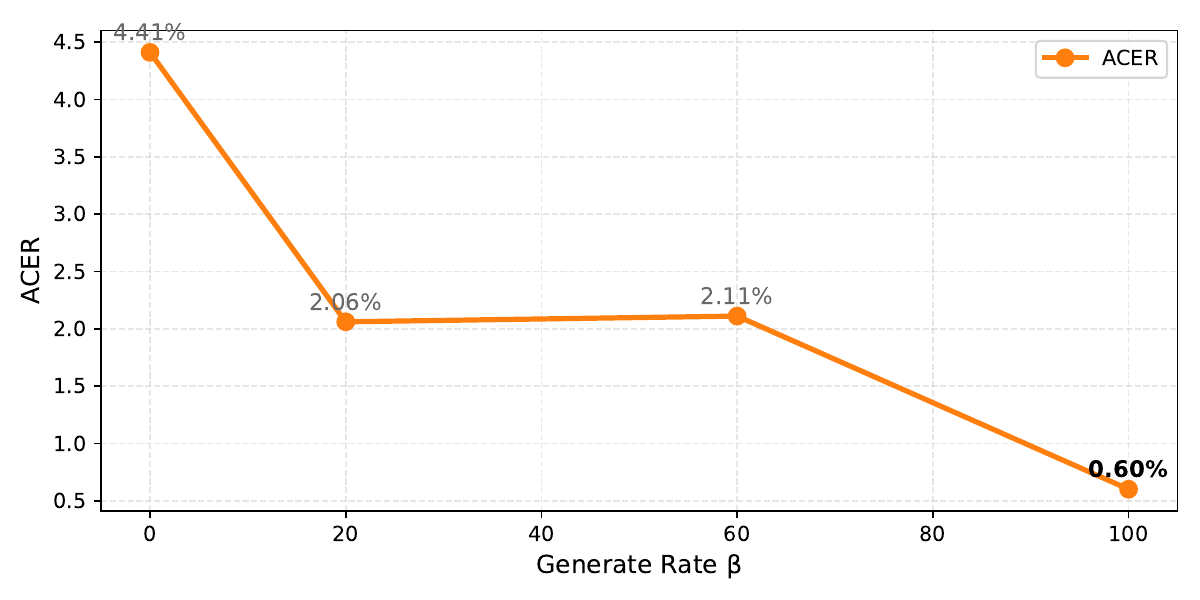}
    \caption{Effect of the test-time completion percentage $\beta$ on WMCA when 70\% of test samples have missing auxiliary modalities. The full detector, including SR-FDA, is fixed. $\beta$ denotes the percentage of incomplete samples whose missing modalities are completed; $\beta=0$ uses only available modalities and $\beta=100$ completes all missing modalities in that subset.}
    \label{fig:ablation_modality_translation}
\end{figure}

To examine the effect of the modality completion ratio, we evaluate a fixed, trained FMA-Net detector including SR-FDA on WMCA in a test scenario where 70\% of samples have missing auxiliary modalities. The completion percentage is defined as $\beta = 100 N_{\mathrm{c}}/N_{\mathrm{m}}$, where $N_{\mathrm{m}}$ is the number of modality-incomplete test samples and $N_{\mathrm{c}}$ counts those receiving translation-based completion of their missing modalities. Thus, $\beta$ is measured relative to the incomplete subset, not the entire test set. At $\beta=0$, the detector uses only the available modalities; at $\beta=100$, all missing modalities in that subset are completed, while acquired observations are retained. Only the test inputs change; the detector is not retrained. Figure~\ref{fig:ablation_modality_translation} shows an overall reduction in ACER as completion increases, with intermediate fluctuations and an ACER of 0.60\% at $\beta=100$. This experiment evaluates the effect of test-time completion under the stated sample-level missingness setting.

\subsection{Visualization Analysis}

\begin{figure}
    \centering
    \includegraphics[width=\linewidth]{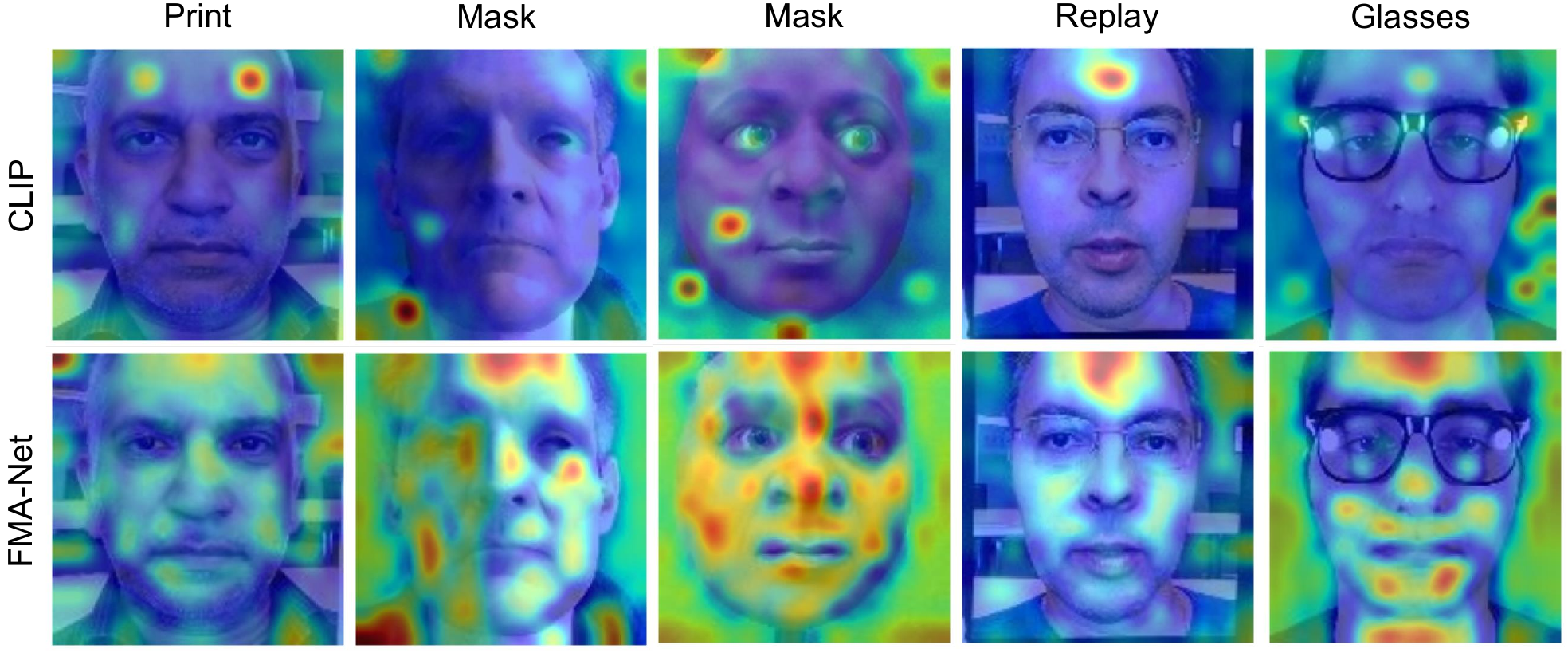}
    \caption{Comparison of CLIP and FMA-Net feature-attention heatmaps. Top is CLIP and bottom is FMA-Net. FMA-Net exploits a richer set of discriminative cues. The source face images are from the WMCA dataset~\cite{george2019biometric}.}
    \label{fig:visulize_hot_feature}
\end{figure}

To further analyze the model's attention areas in handling spoof detection tasks, the Gradient-weighted Class Activation Mapping (Grad-CAM)~\cite{2020Grad} method was used for model visualization and interpretation. Fig.~\ref{fig:visulize_hot_feature} presents the attention heatmaps for two sets of sample images from the WMCA dataset, comparing features extracted by CLIP and the proposed FMA-Net. It can be observed that CLIP tends to allocate attention to irrelevant regions, whereas FMA-Net focuses more effectively on informative areas related to presentation attack cues. This demonstrates that FMA-Net better captures discriminative artifacts and improves detection reliability.

\begin{figure}
    \centering
    \includegraphics[width=\linewidth]{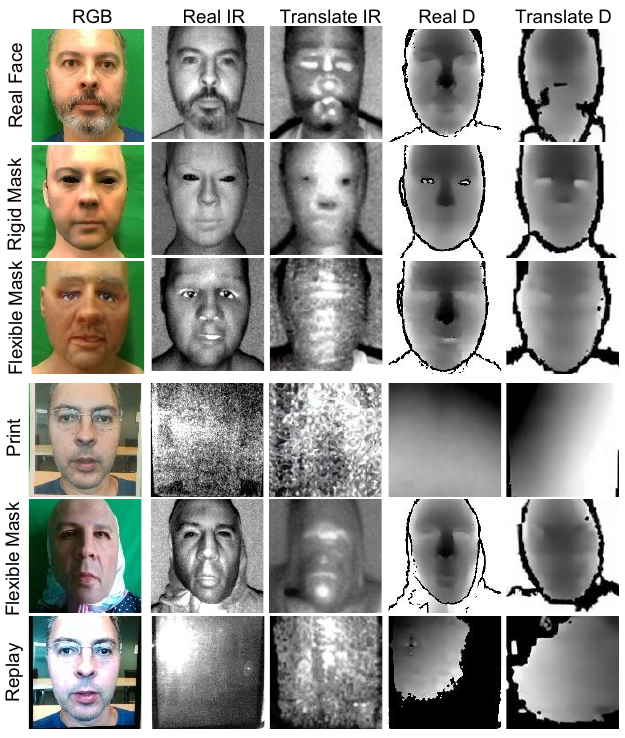}
    \caption{Examples of modality translation. Column 1: ground-truth RGB face images, Columns 2 and 4: corresponding ground-truth IR and depth images, Columns 3 and 5: IR and depth images synthesized by FMIT from the RGB input. The source face images are from the WMCA dataset~\cite{george2019biometric}.}
    \label{fig:ablation_visualize}
\end{figure}

To assess the effect of modality translation completion, Fig.~\ref{fig:ablation_visualize} shows partial results of IR and depth modality translation on the WMCA dataset. The second and fourth columns represent the real IR and depth modalities, respectively, while the third and fifth columns show the corresponding modalities translated by FMIT from RGB inputs. For both bona fide faces and 3D mask attacks, the translated IR modality is broadly consistent with the real IR modality in facial contours and feature positions, but local texture discrepancies remain visible. For print and replay attacks, the translated IR observations also present strong reflection and contour attenuation, which are consistent with some IR-domain spoofing characteristics but may not fully preserve all fine-grained local details. The translated depth modality retains facial depth structures, especially around the eye sockets and cheeks, and captures the missing-depth property associated with 2D attacks. Overall, the visualization suggests that FMIT can recover useful modality-level cues, but the translated observations are not perfect substitutes for physically captured modalities. This limitation may contribute to performance variation under complex attacks, high missing ratios, or cross-domain settings.

\label{subsec:Dataset}

\section{Conclusion}

This paper presented FMA-Net, a flexible-modality framework for multimodal face presentation attack detection under incomplete modality availability. By estimating unavailable modality observations from accessible RGB inputs, FMA-Net enables missing-modality compensation without requiring all physical modalities to be captured at inference time. The proposed framework further integrates face-oriented feature adaptation and synthetic-to-real domain alignment to enhance spoof-discriminative representation learning and reduce the discrepancy between translated and real observations. Experiments on WMCA and PADISI-Face demonstrate that FMA-Net achieves competitive performance under both intra-domain and cross-domain flexible-modality protocols.

The results also reveal several limitations. Translated observations can provide useful complementary cues, but they cannot fully replace physically acquired IR or depth modalities, especially under high missing ratios, large domain shifts, or attacks dominated by fine-grained reflection and material cues. Synthetic-to-real alignment is beneficial in some settings but does not consistently improve cross-dataset generalization, because modality-origin alignment and dataset transfer are distinct objectives. Future work will therefore focus on improving the fidelity and attack-cue consistency of modality translation, developing lightweight translators for deployment, and incorporating interpretable multimodal reasoning to further strengthen robustness in practical biometric security systems.

\bibliographystyle{IEEEtran}
\bibliography{ref}

\end{document}